\documentclass[10pt,twocolumn,letterpaper]{article}

\usepackage{cvpr} 
\usepackage{times}
\usepackage{epsfig}
\usepackage{graphicx}
\usepackage{amsmath}
\usepackage{amssymb}
\usepackage{algorithm}
\usepackage{algpseudocode}
\usepackage{multirow}
\usepackage{booktabs}

\usepackage[pagebackref=true,breaklinks=true,letterpaper=true,colorlinks,bookmarks=false]{hyperref}

\begin{document}

\title{A Quantitative Evaluation of the Expressivity of BMI, Pose and Gender in Body Embeddings for Recognition and Identification}
\author{
Basudha Pal\textsuperscript{*}\quad
Siyuan Huang\textsuperscript{*}\quad
Rama Chellappa\quad \\
Johns Hopkins University, Baltimore, MD, USA\\
{\tt\small \{bpal5, shuan124, rchella4\}@jhu.edu}
}


\maketitle
\thispagestyle{empty}
\begin{abstract}
    Person Re-identification (ReID) systems that match individuals across images or video frames are essential in many real-world applications. However, existing methods are often influenced by attributes such as gender, pose, and body mass index (BMI), which vary in unconstrained settings and raise concerns related to fairness and generalization. To address this, we extend the notion of \textit{expressivity}, defined as the mutual information between learned features and specific attributes, using a secondary neural network to quantify how strongly attributes are encoded. Applying this framework to three ReID models, we find that BMI consistently shows the highest expressivity in the final layers, indicating its dominant role in recognition. In the last attention layer, attributes are ranked as $\text{BMI} > \text{Pitch} > \text{Gender} > \text{Yaw}$, revealing their relative influences in representation learning. Expressivity values also evolve across layers and training epochs, reflecting a dynamic encoding of attributes. These findings demonstrate the central role of body attributes in ReID and establish a principled approach for uncovering attribute driven correlations.
\end{abstract}
\vspace{-4mm}
\section{Introduction}
Deep learning models are trained to learn specific target attributes, but often encode unintended image-related attributes that can adversely affect model performance and fairness. In the domain of biometrics, particularly face recognition, Hill et al.~\cite{hill2019deep} demonstrated that deep networks form identity representations that inherently cluster based on gender. Moreover, these identity embeddings have been shown to encode other latent characteristics such as pose, age, and lighting conditions~\cite{hill2019deep, nagpal2019deep, parde2017face}. The presence of these latent attributes can significantly influence the accuracy of recognition algorithms, as they may inadvertently affect model predictions~\cite{givens2013introduction, lee2014generalizing}. Evaluating these biases or understanding the correlation between attributes and network features require a systematic analysis of how these attributes are embedded and how they influence model behavior. A deeper understanding of these phenomena necessitates investigating how facial or body attributes are encoded in identity representations and how they shape predictive outcomes. In this context, Dhar et al.~\cite{dhar2020attributes} introduced the concept of \textit{expressivity}, for face recognition, a metric that quantifies the relationship between learned network features and specific attributes, thereby enhancing the interpretability of face recognition models. Building on this concept, we extend the framework of expressivity to the domain of person ReID with the goal of evaluating how body-related features are embedded within ReID models trained primarily for recognition.
\newline
\indent Person ReID is a well-established research area with a range of real-world applications, including smart city infrastructure for public safety and traffic management~\cite{behera2020person, khan2024deep} and autonomous driving systems for pedestrian detection and tracking~\cite{camara2020pedestrian, wong2020identifying}. The primary objective of ReID is to accurately match and retrieve pedestrian identities across non-overlapping camera views, varying time frames, and distinct locations, while addressing challenges such as pose variations, appearance diversity, and environmental conditions~\cite{gu2022clothes, gu2019temporal, zheng2015scalable}. Significant progress has been achieved in improving ReID accuracy through the development of deep learning methods, which can broadly be categorized into image-based and video-based approaches. Image-based ReID methods focus on selecting the most distinctive frame and extracting fine-grained spatial features, while video-based approaches aggregate temporal information across multiple frames to produce more robust identity representations. Recent advancements have increasingly combined these approaches, leveraging the strengths of both image-level detail and temporal consistency to achieve state-of-the-art (SoTA) performance. Despite these advancements, most deep learning-based ReID systems are trained to identify individuals based on visual body features, without explicitly learning specific body-related attributes. These models generate identity representations derived from body cues; however, similar to face recognition systems, ReID networks often unintentionally encode additional attributes related to body characteristics. To address this, our work conducts a comprehensive analysis of the attributes correlated with feature embeddings generated by the three SoTA ReID models and compares their robustness. Recently, we have come across the work of Metz et al.\cite{metz2025dissecting}, where they also attempt to identify what information beyond identity is stored in the feature vectors from learned body recognition models. While they used an empirical method by training a logistic regression model to predict gender from image embeddings, it primarily demonstrates the presence of linearly separable attribute information. This method relies on performance metrics from a downstream classifier and does not capture the underlying statistical dependencies among attributes and representations. We adopt an information-theoretic perspective by applying Mutual Information Neural Estimation (MINE) to directly quantify the dependency among attribute variables and deep body recognition features. This allows us to measure how much information about an attribute is encoded in the feature space, regardless of classifier performance. Thus by moving beyond specific prediction and directly analyzing feature–attribute dependencies, our method offers a more reliable and theoretically grounded evaluation of attribute leakage and representational bias understanding. As ReID systems are increasingly deployed in real-world applications, there is a growing demand for explainable and transparent models. Understanding how various attributes are encoded across internal network layers is crucial for interpreting identity predictions and identifying potential sources of algorithmic bias. The following are the conceptual and experimental contributions of our paper:
\begin{itemize}
    \item We present the first investigation into the encoding of body attributes within the layers of a large-scale Vision Transformer (ViT)-based self-supervised foundation model called semReID\cite{huang2023self}. We also compare our results against two other transformer based benchmark models. To enhance the interpretability of large-scale deployable ReID systems, we present a post-hoc framework that explains how internal representations influence identity predictions. This underscores the robustness of our method, despite the inherent complexity of the model and the diversity of the dataset.
    \item In the final attention layer of the SemReID network as well as the other networks, we observe the following order of expressivity for body attributes: \textbf{BMI $>$ Pitch $>$ Gender$>$ Yaw}. This ranking highlights the varying degrees of influence that different attributes have on the network's predictions.
    \item To provide a more comprehensive understanding, we analyze how feature-attribute correlations evolve across different layers and throughout the training process. This layer-wise and temporal analysis offers deeper insights into the embedding of body attributes and their impact on ReID performance.
\end{itemize}

\section{Related Works}
Person ReID aims to match individuals across non-overlapping camera views under challenging conditions such as illumination, clothing, pose, and occlusion~\cite{gu2022clothes, gu2019temporal, huang2019celebrities, zheng2015scalable}. Extensive efforts have addressed this problem across domains like Clothes-Changing ReID (CC-ReID)~\cite{gu2022clothes}, video ReID~\cite{cao2023event, hou2020temporal, yan2020learning, zhang2020multi, wu2022cavit}, unconstrained ReID~\cite{cornett2023expanding, liu2024farsight, nikhal2023weakly, nikhal2024hashreid, zhu2024sharc}, and short-term ReID~\cite{chen2023beyond, zhang2020multi, wang2018learning, zhu2022pass}. Among these, SemReID~\cite{huang2023self} achieves SoTA performance across all four domains. While ReID interpretability remains underexplored, broader recognition systems, especially face recognition have received more attention.

Bias and interpretability in biometrics have long been studied~\cite{schwemmer2020diagnosing, dhar2021pass, siddiqui2022examination, pal2024gamma, pal2024diversinet}. Schumann et al.~\cite{schumann2017person} used an auxiliary network to enrich CNN features, and Myers et al.~\cite{myers2023recognizing} leveraged both linguistic and non-linguistic body representations for identity prediction. These works analyze model sensitivity to attributes via concept-based prediction changes. Yin et al.~\cite{yin2019towards} introduced a spatial activation diversity loss to preserve interpretability in face recognition, while Kim et al.~\cite{kim2014bayesian} proposed a prototype-based generative model. However, as noted in~\cite{kim2018interpretability}, such methods are limited to models trained from scratch and do not generalize to deployed networks. Post-hoc interpretability methods offer alternatives, notably TCAV~\cite{kim2018interpretability}, which measures sensitivity to user-defined concepts via Concept Activation Vectors (CAVs) learned through linear classification. While effective for discrete attributes like color or texture, TCAV struggles with continuous or omnipresent attributes (e.g., BMI, pose), where defining negative examples is difficult. TCAV also requires test images to belong to seen classes, limiting use in open-set scenarios. Other methods include layer-wise linear probes~\cite{alain2016understanding}, influence functions~\cite{koh2017understanding}, and saliency-based approaches~\cite{selvaraju2017grad, chattopadhay2018grad}. For ReID specifically, Chen et al.~\cite{chen2021explainable} proposed a pluggable interpreter that attributes image-pair distances to visual cues but depends on metric distillation and is tailored to CNNs. Saliency maps, while helpful for spatial focus, cannot explain abstract or non-localized attributes. Studies in face recognition have further examined attribute hierarchies. Hill et al.~\cite{hill2019deep} revealed that identity representations are nested under sex, illumination, and viewpoint, while Dhar et al.~\cite{dhar2020attributes} used expressivity-based evaluations to identify a hierarchy where age dominates, followed by sex and yaw.

We propose expressivity as a general framework to assess person ReID systems by quantifying how well an attribute can be predicted from learned features. Unlike prior approaches, expressivity applies to both categorical and continuous attributes and is agnostic to model backbone. We demonstrate its utility using ViT-based ReID models, offering insights into how body-related features are embedded and their impact on model performance, paving the way for more interpretable and explainable ReID systems.
\begin{figure*}[!htbp]
  \centering
  \includegraphics[width=0.8\linewidth]{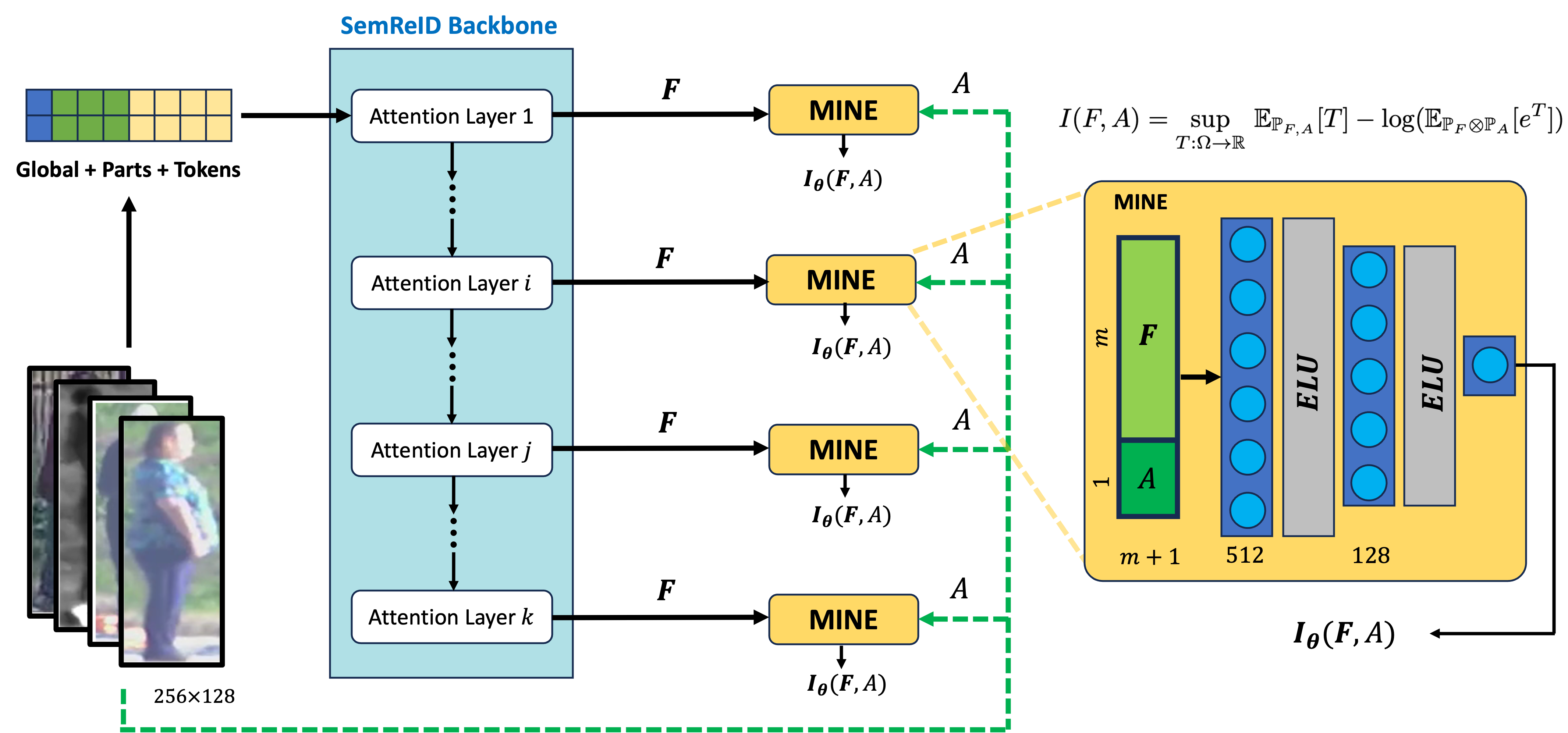}

  \caption{Integrating the MINE block with the ViT based SemReID \cite{huang2023self} backbone to compute the expressivity of features with respect to attributes such as BMI, gender, pitch and yaw. The internal structure of the MINE block employs a simple MLP with two hidden layers to compute the expressivity of \( m \)-dimensional features \( F \). By augmenting these features with an attribute vector \( A \), the input to the network is extended to \( (m+1) \)-dimensions. All subjects involved provided informed consent for their participation, including the use of their images in research publications and figures.}
  \vspace{-3mm}
    \label{mainfig}
 
\end{figure*}
\section{Proposed Method}
Our method as seen in Figure \ref{mainfig} tries to find the correlations between the learnt features by body recognition models and attributes. The predictability of attributes from a given set of body descriptors reflects the amount of attribute-relevant information encoded within those descriptors. To quantify this information, we employ Mutual Information (MI) as shown in Equation~\ref{eqn1}. MI is a fundamental quantity for measuring the relationship between random variables, indicating how much knowledge of one variable reduces uncertainty about the other. By estimating the MI between features learned by the body recognition model and their corresponding sensitive attributes, we assess the degree to which these descriptors encode attribute information. Since MI captures non-linear statistical dependencies between variables and is applicable to both categorical and continuous attributes, this approach provides a unified and consistent measure across attribute types. To develop a general-purpose estimator, we utilize the widely recognized formulation of MI as the Kullback-Leibler (KL) divergence (Kullback, 1997) between the joint distribution and the product of the marginal distributions of two random variables \(X\) and \(Z\), as expressed in Equation~\ref{eqn2}.

\begin{equation}
\label{eqn1}
    I(X ; Z) = \int_{\mathcal{X} \times \mathcal{Z}} \log \frac{d \mathbb{P}_{XZ}}{d \mathbb{P}_X \otimes \mathbb{P}_Z} d \mathbb{P}_{XZ}
\end{equation}

\begin{equation}
\label{eqn2}
    I(X ; Z) = D_{KL}\left(\mathbb{P}_{XZ} \| \mathbb{P}_X \otimes \mathbb{P}_Z\right)
\end{equation}

\subsection{Problem Setup}
\begin{figure}[t]
  \centering
  \includegraphics[width=0.8\linewidth]{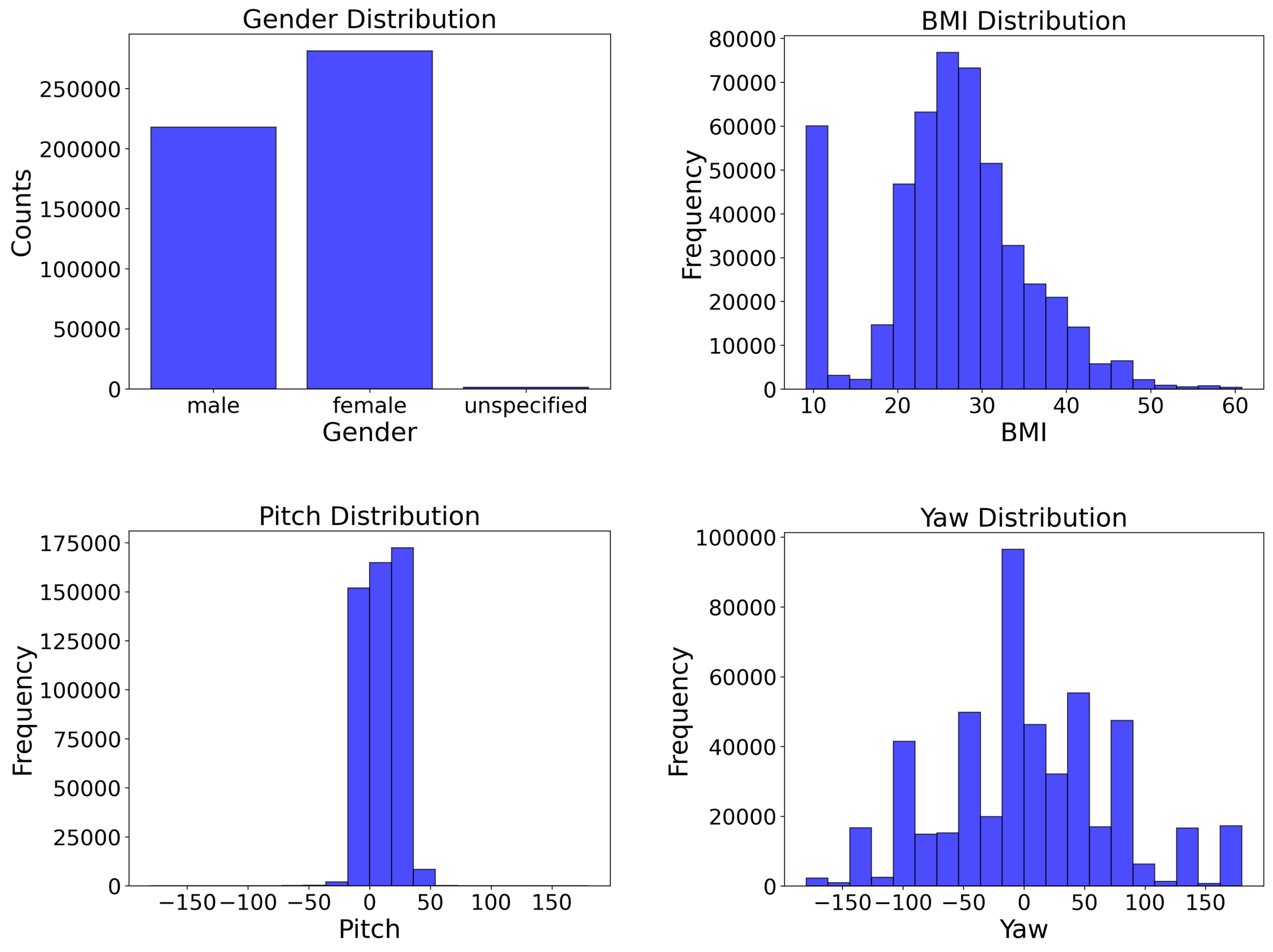}

  \caption{Attribute distribution and counts in the BRIAR dataset indicate sufficient variation across the attributes of interest.}
  \vspace{-4mm}
    \label{attr}

\end{figure}

Our dataset comprises body images of different individuals captured under varying conditions and at different distances. Each image is annotated with an identity label and several sensitive attributes, including gender (\( g \)), height (\( h \)), weight (\( w \)), body mass index (\( BMI \)), which is computed from \( h \) and \( w \), as well as pitch angles (\( p \)) and yaw angles (\( y \)). These attributes collectively form a diverse set of information, enabling a comprehensive analysis of how sensitive attributes are encoded in the learned features. We denote the set of learned feature descriptors as \( \mathbf{F} \), and the corresponding sensitive attributes as \( A \). The primary objective of this analysis is to quantify and explore the correlations between \( \mathbf{F}\) and \( A \) using MINE. Specifically, we aim to estimate the MI, denoted as \( I_\theta(\mathbf{F}, \mathbf{A}) \), to gain insights into how effectively the learned features capture attribute-relevant information. 
To achieve this, each image \( x_i \) undergoes a series of preprocessing steps before being passed through the person ReID models. The model extracts feature descriptors \( f_i \), where \( f_i \in \mathbb{R}^m \), representing the encoded identity and attribute information for each image. These descriptors are then concatenated to form the feature matrix \( \mathbf{F} = [f_1, f_2, \dots, f_n]^T \), where \( F \in \mathbb{R}^{n \times m} \). The sensitive attribute vector \( A \in \mathbb{R}^{n \times 1} \), containing information such as gender, pose, and identity, is then combined with \( F \) to form an augmented matrix \( \mathbf{X} = [\mathbf{F} | \mathbf{A}] \). This augmented matrix \( X \) is subsequently used by the MINE network to estimate the MI between \(\mathbf{F} \) and \( \mathbf{A} \).

MINE employs a neural network-based approach to approximate the MI, enabling us to compute \( I_\theta(\mathbf{F}, \mathbf{A}) \) effectively, even in high-dimensional feature spaces. By leveraging this approach, we can evaluate the extent to which the learned feature descriptors \( \mathbf{F} \) encode information relevant to the sensitive attributes \( \mathbf{A} \). This analysis provides valuable insights into the relationship between the network’s internal representations and sensitive attributes, helping to understand potential biases and attribute-specific influences in the model’s learned features.

\subsection{Attributes and Their Relevance}
We compute the expressivity of four annotated attributes: \(g\), \(BMI\), \(p\) and \(y\) in the extracted  features. In Figure ~\ref{attr}, we verify that the dataset we utilize shows enough variation with respect to these attributes, so that we can ensure that expressivity (which is a lower bound estimate of MI) is an accurate model for the corresponding attributes. When considering g, the vector \(\mathbf{A}\) is a discrete vector having a value of 1 if the gender is male and 0 if female while for \(BMI\), y and p (in degrees) the values of the vector values are continuous.  These attributes play a vital role in person ReID tasks, as they influence the model's learned features. 

\subsection{Expressivity of Body Features}  

Understanding the expressivity of learned features in deep networks is critical for tasks that rely on nuanced feature representations, such as person ReID. Tishby and Zaslavsky~\cite{tishby2015deep} introduced the concept of utilizing MI as a quantitative metric to assess how well information is retained or transformed across the layers of a deep network. MI measures the dependency between random variables, offering insights into the trade-offs between compression and informativeness at various stages of a network. By quantifying MI, one can directly evaluate how effectively the network balances these competing objectives.  
\newline
However, estimating MI for high-dimensional continuous variables is computationally challenging due to the need to compute probability density functions of the underlying distributions. Traditional methods often rely on discretization or kernel density estimation, both of which suffer from scalability issues as dimensionality increases. To overcome this, Belghazi et al.~\cite{belghazi2018mutual} proposed MINE, a scalable framework that approximates MI using a neural network. This bypasses the need for explicit density computations by optimizing a neural network-based lower bound of MI, making it suitable for high-dimensional and complex datasets.  
\newline
The MI between learned feature descriptors \( \mathbf{F} \) and sensitive attributes \( \mathbf{A} \) is a crucial metric in evaluating the expressivity of the learned features. In the context of this work, \( \mathbf{F} \) represents the feature embeddings produced by the ReID model, while \( \mathbf{A} \) denotes associated sensitive attributes such as gender, pose, and identity. The MI approximate is mathematically defined as \(I_{\theta}(\mathbf{F}, \mathbf{A}) = \sup_{\theta \in \Theta} \mathbb{E}_{P_{FA}} \left[ T_{\theta}(f, a) \right] - \log \mathbb{E}_{P_{F} \otimes P_{A}} \left[ e^{T_{\theta}(f)} \right]\), where \( T_{\theta}(f, a) \) is a neural network parameterized by \(\theta\), designed to approximate the MI. The joint expectation \( \mathbb{E}_{P_{FA}} \left[ T_{\theta}(f, a) \right] \) measures the network’s output when conditioned on the true joint distribution of features and attributes. In contrast, the term \( \log \mathbb{E}_{P_{F} \otimes P_{A}} \left[ e^{T_{\theta}(f)} \right] \) normalizes the MI estimate to ensure it captures only the dependency between \( \mathbf{F} \) and \( \mathbf{A} \), excluding any bias from their marginal distributions.  

\subsubsection*{Computational Steps for MI Estimation:}  
\paragraph{Step 1: Joint Expectation Approximation.}  
The first term, \( \mathbb{E}_{P_{FA}} \left[ T_{\theta}(f, a) \right] \), quantifies the degree of dependency between \( \mathbf{F} \) and \( \mathbf{A} \) by evaluating the network output over their joint distribution. In practice, this expectation is approximated over minibatches of data as:  
\vspace{-3mm}
\begin{equation}
\mathbb{E}_{P_{FA}} \left[ T_{\theta}(f, a) \right] \approx \frac{1}{b} \sum_{i=1}^{b} T_{\theta}(f_i, a_i),
\end{equation}  

where \( b \) is the batch size, and \( f_i \) and \( a_i \) are the \( i \)-th feature vector and attribute value in the batch, respectively. This term essentially aggregates the network’s outputs for each feature-attribute pair, capturing their joint statistics.  
\vspace{-3mm}
\paragraph{Step 2: Marginal Expectation Approximation.}  
The second term, \( \log \mathbb{E}_{P_{F} \otimes P_{A}} \left[ e^{T_{\theta}(f)} \right] \), ensures that the MI estimate reflects only the mutual dependency, independent of marginal distributions. It is computed by approximating the expectation of the exponential of the network’s output under the product of marginals:  

\begin{equation}
\mathbb{E}_{P_{F} \otimes P_{A}} \left[ e^{T_{\theta}(f)} \right] \approx \frac{1}{b} \sum_{i=1}^{b} e^{T_{\theta}(f_i)}.
\end{equation}  
This term prevents the MI estimate from over-representing trivial correlations caused by the underlying marginal distributions.  
\vspace{-3mm}
\paragraph{Step 3: Objective Function Formulation.}  
The MI lower bound is approximated as the difference between the joint and marginal expectations, yielding the objective function:  
\vspace{-3mm}
\begin{equation}
V(\theta) = \frac{1}{b} \sum_{i=1}^{b} T_{\theta}(f_i, a_i) - \log \left( \frac{1}{b} \sum_{i=1}^{b} e^{T_{\theta}(f_i)} \right).
\end{equation}  

Maximizing \( V(\theta) \) corresponds to maximizing the MI lower bound, thus enabling the neural network to learn representations that effectively capture the mutual dependency between features and sensitive attributes.  
\vspace{-3mm}
\paragraph{Step 4: Loss Function and Optimization.}  
To train the neural network \( T_{\theta} \), the negative of the objective function is used as the loss:  
\vspace{-1mm}
\begin{equation}
L(\theta) = -V(\theta).
\end{equation}  
\vspace{-1mm}
The gradient of this loss function with respect to the parameters \( \theta \) is computed as: 
\begin{equation}
\nabla_{\theta} L(\theta) = - \left( \mathbb{E}_{P_{FA}} \left[ \nabla_{\theta} T_{\theta} \right] - \mathbb{E}_{P_{F} \otimes P_{A}} \left[ \nabla_{\theta} e^{T_{\theta}} \right] \right).
\end{equation}  

This gradient is then used to iteratively update the network parameters using gradient descent. To mitigate biases introduced by minibatch sampling, an exponential moving average of the gradients is applied during optimization.

In the context of person ReID, this framework is particularly valuable for understanding the expressivity of the feature descriptors generated by the model. The neural network \( T_{\theta} \) is trained to approximate the MI between the learned features \( \mathbf{F} \) and sensitive attributes \(\mathbf{A} \). By iteratively computing joint and marginal expectations and updating \( \theta \), the MI provides a robust metric to quantify how much attribute-relevant information is encoded in the features. At convergence, it reflects the extent to which the model's representations capture sensitive attribute information, offering insights into the expressivity and fairness of the learned features.  
\vspace{-0.5mm}
\section{Experiments}
\subsection{Dataset and Settings}
\begin{figure}[htbp]
  \centering
  \includegraphics[width=0.7\linewidth]{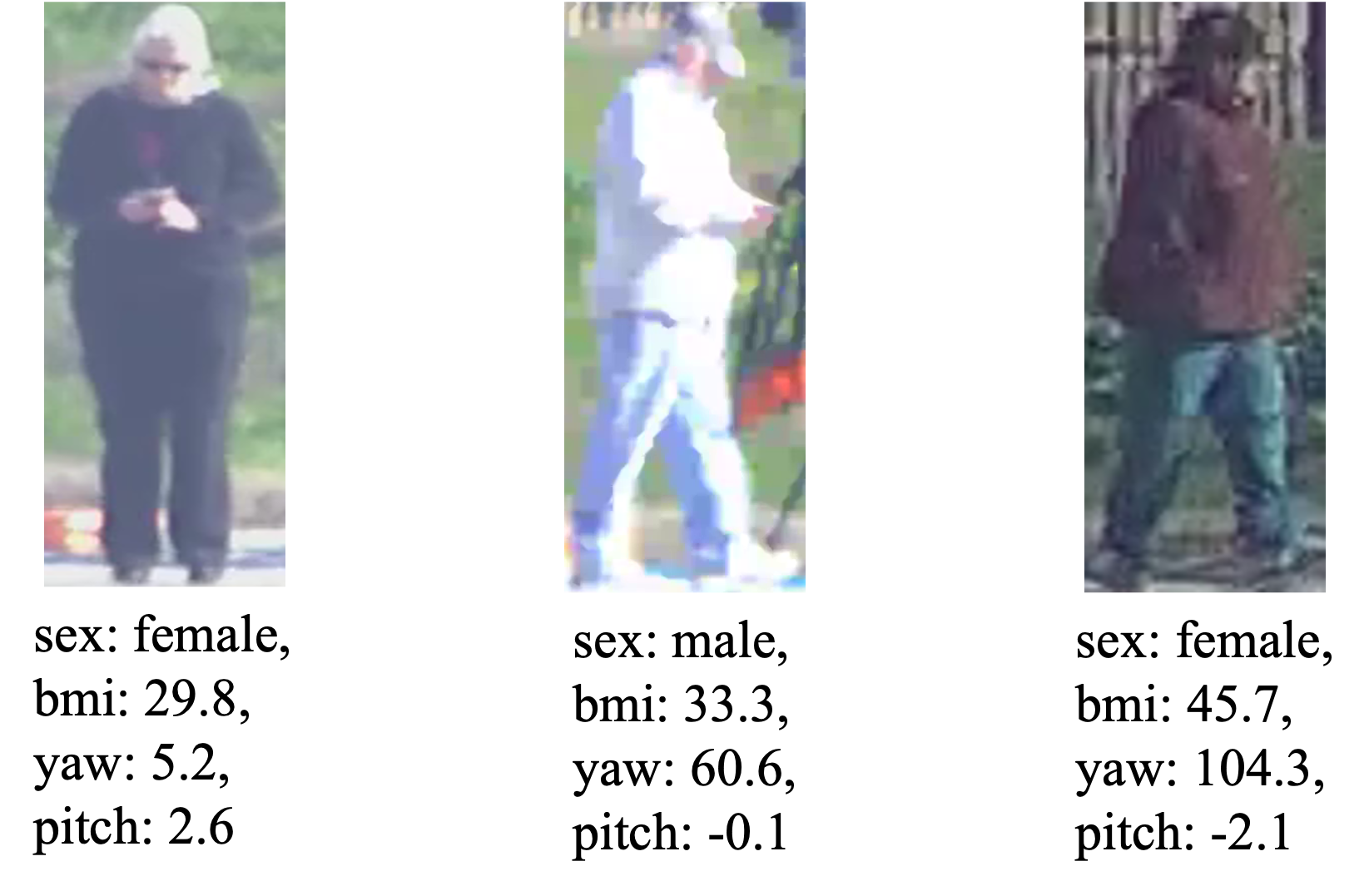}
  \caption{Attribute annotated exemplar images from the BRIAR dataset. All subjects involved provided informed consent for their participation, including the use of their images in research publications and figures.}
  \vspace{-2mm}
  \label{examplesubj}
 
\end{figure}

We use the BRIAR 1–5 dataset~\cite{cornett2023expanding}, a large-scale unconstrained person re-identification benchmark comprising over 1 million images and 40,000 videos captured under real-world conditions, including varying clothing, distances (100m–1km), altitudes (e.g., UAV), and environmental challenges like occlusion, blur, and turbulence. BRIAR includes five progressively complex subsets (BRIAR-1 to 5), increasing in identities, distractors, and capture variability. For our study, we extract 704,999 frames from 382,229 images and 170,522 videos, covering 2,077 unique identities (887 male and 1,190 female subjects). Figure~\ref{examplesubj} shows examples of images and attribute annotations from our curated subset.

\subsection{Integration of MINE with ReID Models}  
\begin{algorithm}[H]
\caption{Expressivity Computation on learnt representations}
\label{alg:expressivity}
\begin{algorithmic}[1]
\Require Layer \( L \), set of \( n \) images \( I \), attribute vector \( \mathbf{A} \in \mathbb{R}^{n \times 1} \)
\Ensure Expressivity measure
\State Initialize \( E \gets [ ] \) \Comment{To store expressivity values}
\State Extract features \( \mathbf{F} \gets [f_1, f_2, \dots, f_n]^T \) from \( L \) after a particular epoch for all \( i \in I \)
\State Concatenate the features and attributes: \( \mathbf{X} \gets [\mathbf{F} | \mathbf{A}] \) \Comment{Augmentation step}
\For{\( \text{iteration} = 1 \) to \( M \)}
    \State Initialize MINE network \( T_\theta \) based on the dimensions of \( \mathbf{X} \)
    \State Compute expressivity score: \( e \gets \text{MINE}(\mathbf{X}) \)
    \State Append score: \( E \gets E \cup \{e\} \)
\EndFor
\State \Return \( \text{Expressivity} \gets \text{Average}(E) \)
\end{algorithmic}
\end{algorithm}
We analyze attribute correlation trends over three SoTA person ReID models namely Pose-guided Feature Disentangling (PFD)\cite{wang2022pose}, Dc-former\cite{li2023dc} and SemReID \cite{huang2023self}. PFD introduces a transformer-based framework for occluded person Re-ID, using pose keypoints to guide feature disentanglement between visible and occluded body regions. It leverages a ViT-base backbone to encode patch-level features and a transformer decoder with learnable semantic tokens aligned to pose-guided regions via a Pose-View Matching module. A pose-guided push loss further encourages discrimination of visible parts while suppressing noisy occluded features for robust identity matching. The DC-Former utilizes a ViT-small backbone to improve person ReID by learning multiple diverse and compact embedding subspaces. It uses multiple class tokens within the Vision Transformer framework to represent distinct embedding spaces. A Self-Diverse Constraint (SDC) is applied to encourage these embeddings to be both diverse and compact, improving the model's ability to distinguish between similar identities. Additionally, a Dynamic Weight Controller (DWC) balances the importance of each embedding space during training, leading to robust and discriminative feature representations. SemReID is a self-supervised person ReID model that introduces a novel Local Semantic Extraction (LSE) module. This module uses keypoint predictions to guide the Segment Anything Model (SAM), producing precise local semantic masks for various body parts. These masks allow for the extraction of fine-grained biometric features, enhancing identity discrimination. SemReID is trained using a teacher-student framework with multiple loss functions to promote robustness and transferability. At inference, only the teacher network and a single linear layer are used, enabling efficient re-identification by processing input data through the teacher encoder. Global and local features are concatenated to facilitate generalization across domains without domain-specific fine-tuning.
\newline 
\indent To evaluate feature expressivity, MINE is integrated into these pipelines as an auxiliary neural network estimating mutual information (MI) by maximizing the Donsker-Varadhan (DV) lower bound. Given a dataset of \( n \) images with corresponding sensitive attributes \( \mathbf{A} \in \mathbb{R}^{n \times 1} \), features are extracted from a specific layer \( L \), yielding \( \mathbf{F} = [f_1, f_2, \dots, f_n]^T \). These features, capturing both global and local cues, are concatenated with attributes to form the input matrix \( \mathbf{X} = [\mathbf{F} \,|\, \mathbf{A}] \). The MINE network \( T_\theta \), a multi-layer perceptron (MLP) with hidden layers of 512 and 128 units and ELU activations, is initialized based on \( \mathbf{X} \)'s dimensions. Over \( M \) iterations, \( T_\theta \) processes \( \mathbf{X} \) to compute expressivity scores \( e \), which estimate the MI between features and attributes. These scores are stored in a list \( E \), and the final expressivity is computed as their average (Algorithm~\ref{alg:expressivity}). This integration provides a principled and scalable approach to quantify attribute-relevant information in learned representations. The iterative MINE process ensures stable and unbiased estimates of expressivity while remaining computationally efficient, as it operates on pre-extracted features from these models.


\subsection{Hierarchical and Temporal Analysis of Attribute Influence}

To comprehensively analyze attribute influence in our framework, we examine feature–attribute correlations both hierarchically across model layers and temporally over training epochs. The SemReID model and the PFD model use a ViT backbone with 12 attention layers, while the Dc-former model uses 8 attention layers to capture rich global and local semantics.
\newline
\noindent
\textbf{Hierarchical Analysis:} We extract features from layers 2, 4, 6, 9, and 12 for the PFD and SemReID models while features from layers 2,3,4,6 and 8 for the Dc-former to study how attribute correlations evolve with network depth. We consider layers 2,4 to be early 6,9 to be mid and 12 to be late for ViT base while 2,3 to be early, 4,6 to be mid and 8 to be late for ViT small. These layers are selected to provide a fine-grained view of the learning process from early layers capturing basic spatial patterns to deeper layers encoding high-level semantics.
\newline
\noindent
\textbf{Temporal Analysis:} We train each of these models for 11 epochs and to assess how these correlations change during training, we analyze the final layer features at epochs 1, 3, 5, 8, and 11. Early epochs (1, 3) highlight the emergence of attribute encoding, while later epochs (8, 11) illustrate how these representations stabilize as the model converges.
\newline
\indent Together, this dual analysis, provides a detailed understanding of how attribute information is processed, encoded, and evolved within these models. It uncovers key trends in the model’s capacity to learn, refine, or suppress sensitive attribute correlations throughout training.
\begin{table}[t]
\centering
\resizebox{0.8\linewidth}{!}{ 
\begin{tabular}{l|ccc|cc}
\toprule
\textbf{Models} & \textbf{Rank 1} & \textbf{5} & \textbf{10} & \textbf{TAR @1\%} & \textbf{@10\%} \\
\midrule
DC-Former & 27.98 & 54.06 & 62.28 & 39.84 & 84.51 \\
PFD       & 32.92 & 55.65 & 75.73 & 47.97 & 71.22 \\
SemReID   & \textbf{34.84} & \textbf{56.95} & \textbf{66.33} & \textbf{54.09} & \textbf{89.03} \\
\bottomrule
\end{tabular}
}
\vspace{0.5mm}
\caption{Quantitative comparison between DC-Former, PFD, and SemReID for identification task (Best in \textbf{bold}).}
\vspace{-4mm}
\label{tab:briar_protocol2}
\end{table}
\subsection{Implementation Details}

We initialize the MINE network based on the input dimensions of the augmented matrix, using a two-layer MLP (512 and 128 units, ELU activations) as seen in Fig. 2 to compute \(T_\theta\). The weights are initialized using Xavier normal initialization. This setup, consistent across experiments, is trained with Adam (learning rate=0.001, batch size=100) until Equation (7) converges. Only the input layer adapts to the feature dimensionality. Expressivity is calculated per Algorithm \ref{alg:expressivity}, with \(M=5\).
For SemReID, we use ViT variants \cite{dosovitskiy2020image} with $384 \times 128$ inputs in a single forward pass. A dual-stream setup extracts 768-dim global and \(3 \times 768\)-dim local semantic features (face, upper, lower body), averaged for the final local embedding. Multi-crop augmentation \cite{caron2020unsupervised,caron2021emerging} uses \(M=2\) global and \(N=3\) local views, followed by \(L=12\) cross-attention layers. Identity embeddings are computed via a BN layer for efficiency. Final 1536-dim features are concatenated with attribute vectors and fed to MINE to estimate MI.
DC-Former \cite{li2023dc} uses multiple class tokens with SDC ($\lambda=1$) to create diverse embedding subspaces, using dynamic weight controller for token balancing.
PFD \cite{wang2022pose} utilizes overlapping patch embedding (step size=12), pose-guided feature aggregation via HRNet-estimated keypoints (threshold $\gamma=0.2$), and part view based decoder with $N_v=17$ learnable semantic views;  decoder layers set to 6.

\begin{table*}[!htbp]
\centering
\scriptsize
\setlength{\tabcolsep}{4pt}
\renewcommand{\arraystretch}{1.2}
\begin{tabular}{l c|ccc|ccc|ccc|ccc}

\toprule
\multicolumn{2}{c|}{} & \multicolumn{3}{c|}{\textbf{Gender}} & \multicolumn{3}{c|}{\textbf{Yaw}} & \multicolumn{3}{c|}{\textbf{BMI}} & \multicolumn{3}{c}{\textbf{Pitch}} \\
\textbf{} & \textbf{No.} & DCFormer & SemReID & PFD & DCFormer & SemReID & PFD & DCFormer & SemReID & PFD & DCFormer & SemReID & PFD \\
\midrule
\multirow{5}{*}{\textbf{Layer}} 
& early (2-s,2-b)     & 0.3252 & 0.0978 & 0.3250 & 0.3109 & 0.0026 & 0.5443 & 0.3588 & 0.0950 & 0.5153 & 0.4430 & 0.1461 & 0.6151 \\
& early (3-s, 4-b)  & 0.3330 & 0.0999 & 0.4176 & 0.3595 & 0.0210 & 0.7181 & 0.4102 & 0.2174 & 0.6509 & 0.4733 & 0.2243 & 0.7259 \\
& mid (4-s, 6-b)      & 0.3490 & 0.1829 & 0.4321 & 0.3458 & 0.0504 & 0.7611 & 0.5023 & 0.2816 & 0.7940 & 0.4203 & 0.2998 & 0.7678 \\
& mid (6-s, 9-b)      & 0.3101 & 0.2150 & 0.4428 & 0.1909 & 0.0716 & 0.5369 & 0.6101 & 0.4714 & 1.2750 & 0.3440 & 0.3190 & 0.5925 \\
& late (8-s,12-b)     & 0.2856 & 0.1586 & 0.4280 & 0.1456 & 0.0119 & 0.4074 & 0.6832 & 0.5740 & 1.2316 & 0.3347 & 0.2727 & 0.4950 \\
\midrule
\multirow{5}{*}{\textbf{Epoch}} 
& 1  & 0.3272 & 0.2115 & 0.4864 & 0.2454 & 0.2519 & 0.7119 & 0.4968 & 0.4283 & 0.9898 & 0.3065 & 0.3722 & 0.7084 \\
& 3  & 0.2980 & 0.1388 & 0.4403 & 0.1772 & 0.000094 & 0.6426 & 0.6198 & 0.3724 & 1.0660 & 0.3332 & 0.2419 & 0.5998 \\
& 5  & 0.2945 & 0.1224 & 0.4425 & 0.1756 & 0.000029 & 0.5512 & 0.6580 & 0.4194 & 1.1495 & 0.3172 & 0.2052 & 0.6097 \\
& 8  & 0.2914 & 0.1461 & 0.4211 & 0.1896 & 0.00145 & 0.4328 & 0.7047 & 0.5479 & 1.2280 & 0.3342 & 0.2517 & 0.5251 \\
& 11 & 0.2856 & 0.1586 & 0.4280 & 0.1456 & 0.0119 & 0.4074 & 0.6832 & 0.5740 & 1.2316 & 0.3347 & 0.2727 & 0.4950 \\
\bottomrule
\end{tabular}
\vspace{0.5mm}
\caption{Expressivity scores for \textbf{Gender}, \textbf{Yaw}, \textbf{BMI}, and \textbf{Pitch} across layers and training epochs for DCFormer, SemReID, and PFD. x-s and y-b denote the $x^{th}$ and $y^{th}$ layer of ViT small and base respectively.}
\vspace{-1mm}
\label{tab:expressivity_booktabs}
\end{table*}
\section{Results and Discussions}
On the person ReID task for BRIAR, SemReID demonstrates the strongest performance (Table \ref{tab:briar_protocol2}). In this section, we analyze feature-attribute correlations using MINE. The discussion is structured into three subsections: (1) spatial trends across the hierarchical feedforward pipeline, (2) temporal progression during training, and finally (3) the advantages of our method.



\subsection{Hierarchical Feedforward Progression of Attribute Expressivity}
\label{sec:hierarchical_analysis}
\begin{figure}[htbp]
  \centering
  \includegraphics[width=\linewidth]{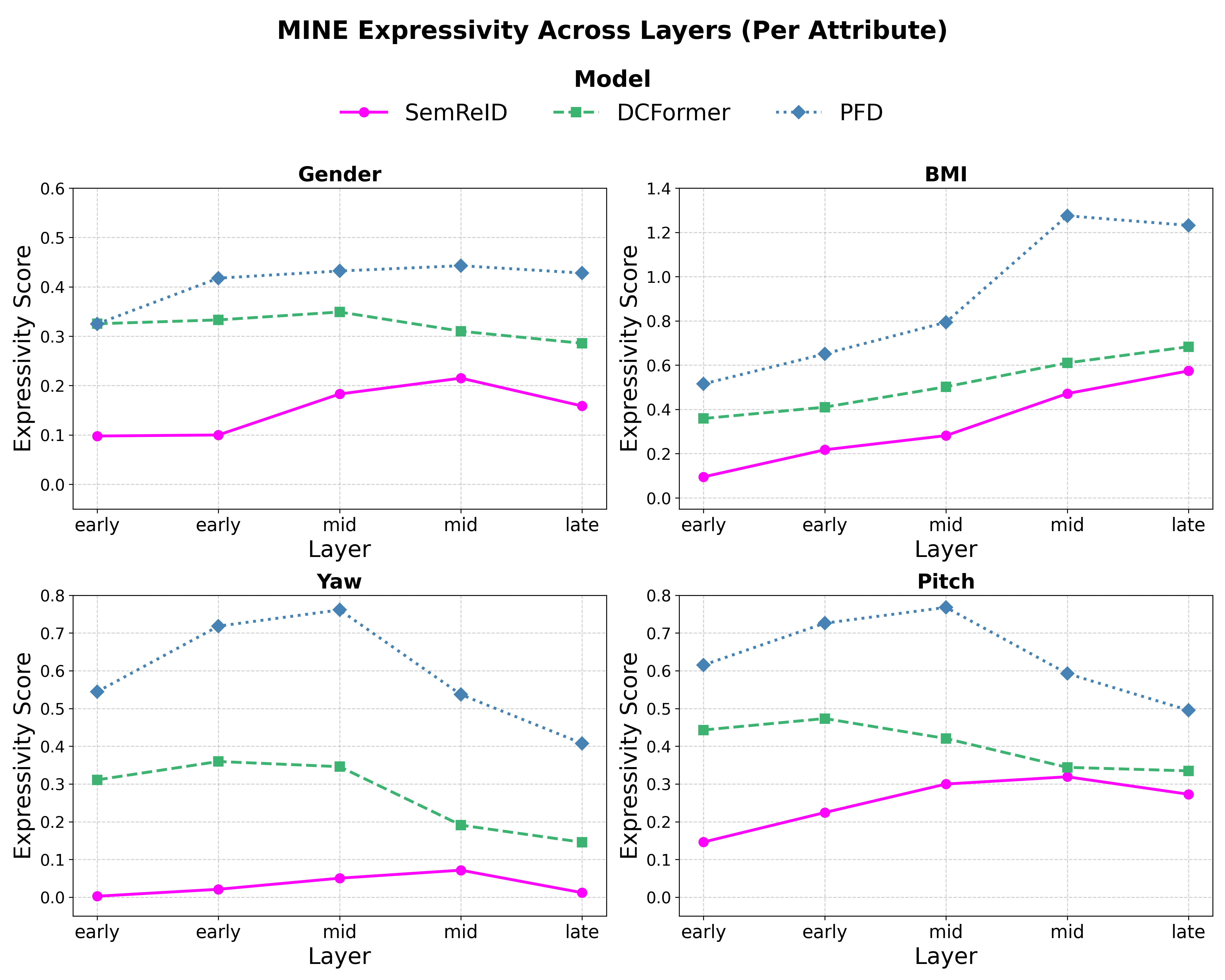}
  \caption{Expressivity trends of gender, yaw, pitch and BMI in input image over layer-wise learnt features from SemReID.}
  \vspace{-1mm}
  \label{layerfig}
 
\end{figure}

To study the evolution of attribute correlations across network depth, we extract features from layers 2, 4, 6, 9, and 12 for the PFD and SemReID models, and from layers 2, 3, 4, 6, and 8 for DCFormer. For ViT-Base (PFD and SemReID), we categorize layers 2 and 4 as \textit{early}, 6 and 9 as \textit{mid-level}, and 12 as \textit{late}. For ViT-Small (DCFormer), layers 2 and 3 are considered \textit{early}, 4 and 6 as \textit{mid-level}, and 8 as \textit{late}. Figure~\ref{layerfig} visualizes attribute expressivity per layer, and Table~\ref{tab:expressivity_booktabs} reports corresponding values. Our analysis reveals the following trends:

\begin{itemize}
    \item \textbf{BMI Encodes Deeply Across All Models:} BMI expressivity increases consistently with depth in all models. PFD shows the sharpest rise from around 0.6 in early layers to over 1.2 in the final block—highlighting that BMI becomes increasingly entangled in deeper feature hierarchies. This suggests that morphometric cues are emphasized more in later layers.
    \vspace{-3mm}

    \item \textbf{Yaw Peaks in Mid-Layers and Then Drops:} Yaw exhibits a peak in the mid-layers and declines in the final block across all models. This is most prominent in PFD, where yaw rises to \(\sim0.75\) mid-network and then drops to \(\sim0.4\). The pattern suggests intermediate layers encode pose cues strongly, which are later partially suppressed as identity features dominate.
    \vspace{-3mm}

    \item \textbf{Pitch Follows a Similar Pattern to Yaw, But Less Drastic:} Pitch also peaks at the mid-layer stage before dropping in the late layers. However, unlike yaw, pitch maintains moderate expressivity even in deeper layers, particularly in DCFormer and PFD. This highlights partial retention of pitch as a latent factor.
    \vspace{-3mm}

    \item \textbf{Gender Expressivity is Moderate and Stable:} Gender shows moderate expressivity across all layers, particularly in PFD and DCFormer, where its scores remain relatively consistent. While it does not peak as strongly as BMI or pose attributes, it is clearly encoded and not suppressed. In SemReID, gender shows a slight rise in mid-layers before declining in the final block. Overall, gender expressivity is less depth-sensitive but remains a persistent latent factor across the network hierarchy.
    \vspace{-3mm}

    \item \textbf{Model-wise Trends:} PFD encodes all attributes more strongly across all layers, indicating greater attribute leakage and less disentanglement. DCFormer exhibits moderate levels of expressivity and clear hierarchical trends. SemReID consistently suppresses attributes more effectively, especially yaw and gender, suggesting more robust identity representation.
\end{itemize}
\vspace{-3mm}
Overall, the feedforward hierarchy reveals that attributes like yaw and pitch are most prominent in intermediate layers, while BMI accumulates in deeper blocks.  In the final layers of all models, we observe a consistent expressivity ranking: \(\text{BMI} > \text{pitch} > \text{gender} > \text{yaw}\).

\subsection{Temporal Progression of Attribute Expressivity}
\label{sec:temporal_analysis}
\begin{figure}[htbp]
  \centering
  \includegraphics[width=\linewidth]{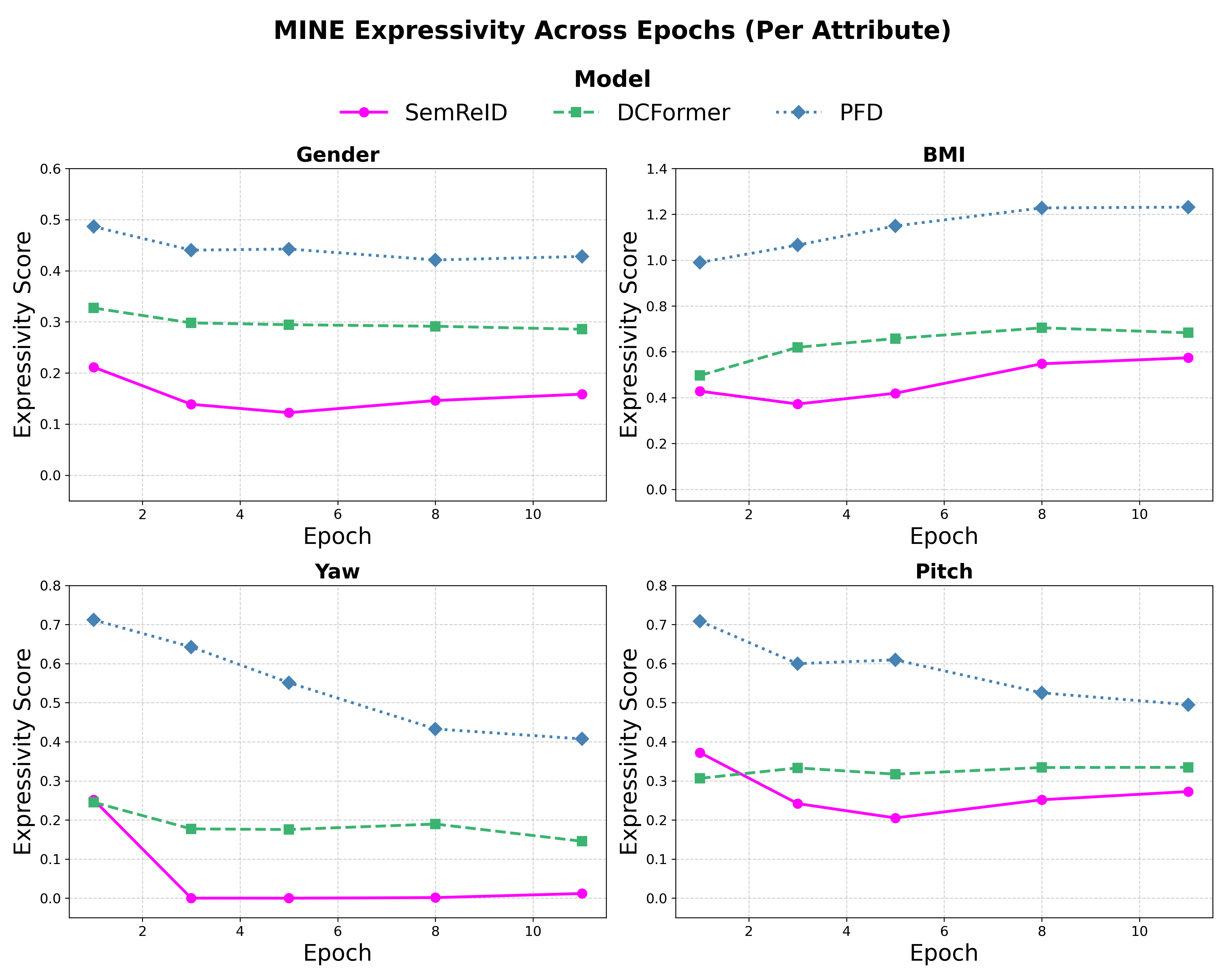}
  \caption{Expressivity trends of gender, yaw, pitch and BMI in input image over epoch-wise learnt features from SemReID.}
  \vspace{-3mm}
  \label{epochfig}

\end{figure}
We analyze the temporal evolution of MINE expressivity scores across epochs for the same four attributes measured at the final attention layer of the three models: SemReID, DCFormer, and PFD. Figure~\ref{epochfig} shows the epoch-wise trends, and Table~\ref{tab:expressivity_booktabs} reports the absolute scores. We observe the following key patterns:

\begin{itemize}
    \item \textbf{Final Attribute Ranking is Consistent Across Models:} By epoch 11, all models exhibit a consistent ranking of attribute expressivity: \(\text{BMI} > \text{pitch} > \text{gender} > \text{yaw}\). This implies that BMI is the most persistently entangled attribute in learned features, while yaw is effectively suppressed over training.
    \vspace{-3mm}
    
    \item \textbf{Yaw Suppression Over Time:} All models exhibit a notable decline in yaw expressivity with training, dropping from over 0.7 to around 0.4 by epoch 11 in PFD, and collapsing to near zero after epoch 2 in SemReID. This indicates active suppression of yaw as models optimize for pose invariant identity representation.
    \vspace{-3mm}
    
    \item \textbf{BMI Persistence Across Models:} Unlike other attributes, BMI expressivity shows a rising or stable trend for all models. In PFD, it increases from 1.0 to 1.2, while DCFormer and SemReID show comparatively modest gains. This highlights BMI as a persistent entangled attribute, potentially acting as a confounding factor in identity representations.
    \vspace{-3mm}
    
    \item \textbf{Gender Expressivity is Low and Stable:} Gender shows the lowest expressivity throughout training across models except for yaw. In SemReID, it drops early and stabilizes near 0.13, indicating progressive decoupling over time.
    \vspace{-3mm}
    
    \item \textbf{Pitch Shows Intermediate Trends:} Pitch expressivity stabilizes after epoch 3 for all models, with DCFormer consistently encoding more pitch information than SemReID. The stability suggests that pitch is preserved during training but is less dominant than BMI.
    \vspace{-7mm}
    \item \textbf{Model Comparisons:} PFD consistently encodes more attribute information than the other two models across all attributes and epochs. In contrast, SemReID exhibits the lowest attribute expressivity overall, indicating more robust identity feature learning with reduced sensitivity to spurious cues like yaw and gender.
\end{itemize}
\vspace{-3mm}
\indent These findings reveal that during training, pose-related cues like yaw are actively suppressed, particularly in SemReID. However, body-morphometric attributes such as BMI persist in the feature space, suggesting a need for targeted disentanglement strategies for such persistent confounders.


    
    


\subsection{Advantages of Expressivity for Person ReID}
In this subsection we further justify the usage of MINE over other existing methods. The key reasons are:
\begin{enumerate}

    \item \textbf{Supports Both Discrete and Continuous Attributes}: Expressivity is versatile and applicable to both discrete (e.g., gender) and continuous (e.g., pitch angle) attributes. For example, gender expressivity can be computed using a binary attribute vector \( A \). Unlike TCAV~\cite{kim2018interpretability}, which is tailored to discrete attributes, expressivity naturally extends to continuous concepts such as pose or BMI where defining clear negative examples is difficult. This makes it especially valuable in ReID, where continuous attributes are often crucial.
\vspace{-7mm}
    \item \textbf{Independent of Training Identities}: Previous methods require computing changes in logits, which limits their applicability to images belonging to training identities. In contrast, expressivity does not rely on logit changes or training identity classes. This independence makes it an effective tool for analyzing unseen attributes not explicitly included during training.
\end{enumerate}

\section{Conclusion}
We propose a method to quantify the information a ViT-based person ReID network learns about various attributes without being explicitly trained on them, by analyzing their expressivity on learnt features. This enables us to identify attributes most relevant to identity recognition across hierarchical layers and training epochs. Several important findings emerge from our investigation: (1) BMI consistently shows the highest expressivity, especially in deeper layers (e.g., layer 12) and later training stages (e.g., epoch 11), making it the most critical attribute for identity recognition even without explicit labels. (2) Attributes like yaw and pitch are expressive in mid-layers (e.g., layers 4 and 6) but lose influence in deeper layers. (3) Temporally, BMI expressivity increases throughout training, while yaw and pitch decline sharply, with yaw showing the steepest drop. Gender is entangled in the features moderately and has minimal variation with the evolution of learned features. These findings highlight BMI as the most significant attribute, followed by yaw gender and pitch for the person ReID task. However, since expressivity approximates MI, it is influenced by entropy and attribute label distribution, potentially affecting cross-attribute comparisons which is an inherent limitation of all MI-based approaches.
\section{Acknowledgements}
SH and RC are supported by the BRIAR project. This research is based upon work supported in part by the Office of the Director of National Intelligence (ODNI), Intelligence Advanced Research Projects Activity (IARPA), via [2022-21102100005]. The views and conclusions contained herein are those of the authors and should not be interpreted as necessarily representing the official policies, either expressed or implied, of ODNI, IARPA, or the U. S. Government.
The US. Government is authorized to reproduce and distribute reprints for governmental purposes notwithstanding any copyright annotation therein.

{\small
\bibliographystyle{ieee}
\bibliography{egbib}

\begin{thebibliography}{10}\itemsep=-1pt

\bibitem{alain2016understanding}
G.~Alain.
\newblock Understanding intermediate layers using linear classifier probes.
\newblock {\em arXiv preprint arXiv:1610.01644}, 2016.

\bibitem{behera2020person}
N.~K.~S. Behera, P.~K. Sa, and S.~Bakshi.
\newblock Person re-identification for smart cities: State-of-the-art and the path ahead.
\newblock {\em Pattern Recognition Letters}, 138:282--289, 2020.

\bibitem{belghazi2018mutual}
M.~I. Belghazi, A.~Baratin, S.~Rajeshwar, S.~Ozair, Y.~Bengio, A.~Courville, and D.~Hjelm.
\newblock Mutual information neural estimation.
\newblock In {\em International conference on machine learning}, pages 531--540. PMLR, 2018.

\bibitem{camara2020pedestrian}
F.~Camara, N.~Bellotto, S.~Cosar, F.~Weber, D.~Nathanael, M.~Althoff, J.~Wu, J.~Ruenz, A.~Dietrich, G.~Markkula, et~al.
\newblock Pedestrian models for autonomous driving part ii: high-level models of human behavior.
\newblock {\em IEEE Transactions on Intelligent Transportation Systems}, 22(9):5453--5472, 2020.

\bibitem{cao2023event}
C.~Cao, X.~Fu, H.~Liu, Y.~Huang, K.~Wang, J.~Luo, and Z.-J. Zha.
\newblock Event-guided person re-identification via sparse-dense complementary learning.
\newblock In {\em Proceedings of the IEEE/CVF Conference on Computer Vision and Pattern Recognition}, pages 17990--17999, 2023.

\bibitem{caron2020unsupervised}
M.~Caron, I.~Misra, J.~Mairal, P.~Goyal, P.~Bojanowski, and A.~Joulin.
\newblock Unsupervised learning of visual features by contrasting cluster assignments.
\newblock {\em Advances in neural information processing systems}, 33:9912--9924, 2020.

\bibitem{caron2021emerging}
M.~Caron, H.~Touvron, I.~Misra, H.~J{\'e}gou, J.~Mairal, P.~Bojanowski, and A.~Joulin.
\newblock Emerging properties in self-supervised vision transformers.
\newblock In {\em Proceedings of the IEEE/CVF international conference on computer vision}, pages 9650--9660, 2021.

\bibitem{chattopadhay2018grad}
A.~Chattopadhay, A.~Sarkar, P.~Howlader, and V.~N. Balasubramanian.
\newblock Grad-cam++: Generalized gradient-based visual explanations for deep convolutional networks.
\newblock In {\em 2018 IEEE winter conference on applications of computer vision (WACV)}, pages 839--847. IEEE, 2018.

\bibitem{chen2023beyond}
W.~Chen, X.~Xu, J.~Jia, H.~Luo, Y.~Wang, F.~Wang, R.~Jin, and X.~Sun.
\newblock Beyond appearance: a semantic controllable self-supervised learning framework for human-centric visual tasks.
\newblock In {\em Proceedings of the IEEE/CVF conference on computer vision and pattern recognition}, pages 15050--15061, 2023.

\bibitem{chen2021explainable}
X.~Chen, X.~Liu, W.~Liu, X.-P. Zhang, Y.~Zhang, and T.~Mei.
\newblock Explainable person re-identification with attribute-guided metric distillation.
\newblock In {\em Proceedings of the IEEE/CVF international conference on computer vision}, pages 11813--11822, 2021.

\bibitem{cornett2023expanding}
D.~Cornett, J.~Brogan, N.~Barber, D.~Aykac, S.~Baird, N.~Burchfield, C.~Dukes, A.~Duncan, R.~Ferrell, J.~Goddard, et~al.
\newblock Expanding accurate person recognition to new altitudes and ranges: The briar dataset.
\newblock In {\em Proceedings of the IEEE/CVF Winter Conference on Applications of Computer Vision}, pages 593--602, 2023.

\bibitem{dhar2020attributes}
P.~Dhar, A.~Bansal, C.~D. Castillo, J.~Gleason, P.~J. Phillips, and R.~Chellappa.
\newblock How are attributes expressed in face dcnns?
\newblock In {\em 2020 15th IEEE International Conference on Automatic Face and Gesture Recognition (FG 2020)}, pages 85--92. IEEE, 2020.

\bibitem{dhar2021pass}
P.~Dhar, J.~Gleason, A.~Roy, C.~D. Castillo, and R.~Chellappa.
\newblock Pass: protected attribute suppression system for mitigating bias in face recognition.
\newblock In {\em Proceedings of the IEEE/CVF International Conference on Computer Vision}, pages 15087--15096, 2021.

\bibitem{dosovitskiy2020image}
A.~Dosovitskiy, L.~Beyer, A.~Kolesnikov, D.~Weissenborn, X.~Zhai, T.~Unterthiner, M.~Dehghani, M.~Minderer, G.~Heigold, S.~Gelly, et~al.
\newblock An image is worth 16x16 words.
\newblock {\em arXiv preprint arXiv:2010.11929}, 7, 2020.

\bibitem{givens2013introduction}
G.~H. Givens, J.~R. Beveridge, P.~J. Phillips, B.~Draper, Y.~M. Lui, and D.~Bolme.
\newblock Introduction to face recognition and evaluation of algorithm performance.
\newblock {\em Computational Statistics \& Data Analysis}, 67:236--247, 2013.

\bibitem{gu2022clothes}
X.~Gu, H.~Chang, B.~Ma, S.~Bai, S.~Shan, and X.~Chen.
\newblock Clothes-changing person re-identification with rgb modality only.
\newblock In {\em Proceedings of the IEEE/CVF conference on computer vision and pattern recognition}, pages 1060--1069, 2022.

\bibitem{gu2019temporal}
X.~Gu, B.~Ma, H.~Chang, S.~Shan, and X.~Chen.
\newblock Temporal knowledge propagation for image-to-video person re-identification.
\newblock In {\em Proceedings of the IEEE/CVF international conference on computer vision}, pages 9647--9656, 2019.

\bibitem{hill2019deep}
M.~Q. Hill, C.~J. Parde, C.~D. Castillo, Y.~I. Colon, R.~Ranjan, J.-C. Chen, V.~Blanz, and A.~J. O’Toole.
\newblock Deep convolutional neural networks in the face of caricature.
\newblock {\em Nature Machine Intelligence}, 1(11):522--529, 2019.

\bibitem{hou2020temporal}
R.~Hou, H.~Chang, B.~Ma, S.~Shan, and X.~Chen.
\newblock Temporal complementary learning for video person re-identification.
\newblock In {\em Computer Vision--ECCV 2020: 16th European Conference, Glasgow, UK, August 23--28, 2020, Proceedings, Part XXV 16}, pages 388--405. Springer, 2020.

\bibitem{huang2023self}
S.~Huang, Y.~Zhou, R.~Prabhakar, X.~Liu, Y.~Guo, H.~Yi, C.~Peng, R.~Chellappa, and C.~P. Lau.
\newblock Self-supervised learning of whole and component-based semantic representations for person re-identification.
\newblock {\em arXiv preprint arXiv:2311.17074}, 2023.

\bibitem{huang2019celebrities}
Y.~Huang, Q.~Wu, J.~Xu, and Y.~Zhong.
\newblock Celebrities-reid: A benchmark for clothes variation in long-term person re-identification.
\newblock In {\em 2019 International Joint Conference on Neural Networks (IJCNN)}, pages 1--8. IEEE, 2019.

\bibitem{khan2024deep}
S.~U. Khan, T.~Hussain, A.~Ullah, and S.~W. Baik.
\newblock Deep-reid: Deep features and autoencoder assisted image patching strategy for person re-identification in smart cities surveillance.
\newblock {\em Multimedia Tools and Applications}, 83(5):15079--15100, 2024.

\bibitem{kim2014bayesian}
B.~Kim, C.~Rudin, and J.~A. Shah.
\newblock The bayesian case model: A generative approach for case-based reasoning and prototype classification.
\newblock {\em Advances in neural information processing systems}, 27, 2014.

\bibitem{kim2018interpretability}
B.~Kim, M.~Wattenberg, J.~Gilmer, C.~Cai, J.~Wexler, F.~Viegas, et~al.
\newblock Interpretability beyond feature attribution: Quantitative testing with concept activation vectors (tcav).
\newblock In {\em International conference on machine learning}, pages 2668--2677. PMLR, 2018.

\bibitem{koh2017understanding}
P.~W. Koh and P.~Liang.
\newblock Understanding black-box predictions via influence functions.
\newblock In {\em International conference on machine learning}, pages 1885--1894. PMLR, 2017.

\bibitem{lee2014generalizing}
Y.~Lee, P.~J. Phillips, J.~J. Filliben, J.~R. Beveridge, and H.~Zhang.
\newblock Generalizing face quality and factor measures to video.
\newblock In {\em IEEE International Joint Conference on Biometrics}, pages 1--8. IEEE, 2014.

\bibitem{li2023dc}
W.~Li, C.~Zou, M.~Wang, F.~Xu, J.~Zhao, R.~Zheng, Y.~Cheng, and W.~Chu.
\newblock Dc-former: Diverse and compact transformer for person re-identification.
\newblock In {\em Proceedings of the AAAI Conference on Artificial Intelligence}, volume~37, pages 1415--1423, 2023.

\bibitem{liu2024farsight}
F.~Liu, R.~Ashbaugh, N.~Chimitt, N.~Hassan, A.~Hassani, A.~Jaiswal, M.~Kim, Z.~Mao, C.~Perry, Z.~Ren, et~al.
\newblock Farsight: A physics-driven whole-body biometric system at large distance and altitude.
\newblock In {\em Proceedings of the IEEE/CVF Winter Conference on Applications of Computer Vision}, pages 6227--6236, 2024.

\bibitem{metz2025dissecting}
T.~M. Metz, M.~Q. Hill, B.~Myers, V.~N. Gandi, R.~Chilakapati, and A.~J. O'Toole.
\newblock Dissecting human body representations in deep networks trained for person identification.
\newblock {\em arXiv preprint arXiv:2502.15934}, 2025.

\bibitem{myers2023recognizing}
B.~A. Myers, L.~Jaggernauth, T.~M. Metz, M.~Q. Hill, V.~N. Gandi, C.~D. Castillo, and A.~J. O’Toole.
\newblock Recognizing people by body shape using deep networks of images and words.
\newblock In {\em 2023 IEEE International Joint Conference on Biometrics (IJCB)}, pages 1--8. IEEE, 2023.

\bibitem{nagpal2019deep}
S.~Nagpal, M.~Singh, R.~Singh, and M.~Vatsa.
\newblock Deep learning for face recognition: Pride or prejudiced?
\newblock {\em arXiv preprint arXiv:1904.01219}, 2019.

\bibitem{nikhal2024hashreid}
K.~Nikhal, Y.~Ma, S.~S. Bhattacharyya, and B.~S. Riggan.
\newblock Hashreid: Dynamic network with binary codes for efficient person re-identification.
\newblock In {\em Proceedings of the IEEE/CVF Winter Conference on Applications of Computer Vision}, pages 6046--6055, 2024.

\bibitem{nikhal2023weakly}
K.~Nikhal and B.~S. Riggan.
\newblock Weakly supervised face and whole body recognition in turbulent environments.
\newblock In {\em 2023 IEEE International Joint Conference on Biometrics (IJCB)}, pages 1--10. IEEE, 2023.

\bibitem{pal2024gamma}
B.~Pal, A.~Kannan, R.~P. Kathirvel, A.~J. O’Toole, and R.~Chellappa.
\newblock Gamma-face: Gaussian mixture models amend diffusion models for bias mitigation in face images.
\newblock In {\em European Conference on Computer Vision}, pages 471--488. Springer, 2024.

\bibitem{pal2024diversinet}
B.~Pal, A.~Roy, R.~P. Kathirvel, A.~J. O’Toole, and R.~Chellappa.
\newblock Diversinet: Mitigating bias in deep classification networks across sensitive attributes through diffusion-generated data.
\newblock In {\em 2024 IEEE International Joint Conference on Biometrics (IJCB)}, pages 1--10. IEEE, 2024.

\bibitem{parde2017face}
C.~J. Parde, C.~Castillo, M.~Q. Hill, Y.~I. Colon, S.~Sankaranarayanan, J.-C. Chen, and A.~J. O’Toole.
\newblock Face and image representation in deep cnn features.
\newblock In {\em 2017 12th ieee international conference on automatic face \& gesture recognition (fg 2017)}, pages 673--680. IEEE, 2017.

\bibitem{schumann2017person}
A.~Schumann and R.~Stiefelhagen.
\newblock Person re-identification by deep learning attribute-complementary information.
\newblock In {\em Proceedings of the IEEE conference on computer vision and pattern recognition workshops}, pages 20--28, 2017.

\bibitem{schwemmer2020diagnosing}
C.~Schwemmer, C.~Knight, E.~D. Bello-Pardo, S.~Oklobdzija, M.~Schoonvelde, and J.~W. Lockhart.
\newblock Diagnosing gender bias in image recognition systems.
\newblock {\em Socius}, 6:2378023120967171, 2020.

\bibitem{selvaraju2017grad}
R.~R. Selvaraju, M.~Cogswell, A.~Das, R.~Vedantam, D.~Parikh, and D.~Batra.
\newblock Grad-cam: Visual explanations from deep networks via gradient-based localization.
\newblock In {\em Proceedings of the IEEE international conference on computer vision}, pages 618--626, 2017.

\bibitem{siddiqui2022examination}
H.~Siddiqui, A.~Rattani, K.~Ricanek, and T.~Hill.
\newblock An examination of bias of facial analysis based bmi prediction models.
\newblock In {\em Proceedings of the IEEE/CVF Conference on Computer Vision and Pattern Recognition}, pages 2926--2935, 2022.

\bibitem{tishby2015deep}
N.~Tishby and N.~Zaslavsky.
\newblock Deep learning and the information bottleneck principle.
\newblock In {\em 2015 ieee information theory workshop (itw)}, pages 1--5. IEEE, 2015.

\bibitem{wang2018learning}
G.~Wang, Y.~Yuan, X.~Chen, J.~Li, and X.~Zhou.
\newblock Learning discriminative features with multiple granularities for person re-identification.
\newblock In {\em Proceedings of the 26th ACM international conference on Multimedia}, pages 274--282, 2018.

\bibitem{wang2022pose}
T.~Wang, H.~Liu, P.~Song, T.~Guo, and W.~Shi.
\newblock Pose-guided feature disentangling for occluded person re-identification based on transformer.
\newblock In {\em Proceedings of the AAAI conference on artificial intelligence}, volume~36, pages 2540--2549, 2022.

\bibitem{wong2020identifying}
K.~Wong, S.~Wang, M.~Ren, M.~Liang, and R.~Urtasun.
\newblock Identifying unknown instances for autonomous driving.
\newblock In {\em Conference on Robot Learning}, pages 384--393. PMLR, 2020.

\bibitem{wu2022cavit}
J.~Wu, L.~He, W.~Liu, Y.~Yang, Z.~Lei, T.~Mei, and S.~Z. Li.
\newblock Cavit: Contextual alignment vision transformer for video object re-identification.
\newblock In {\em European Conference on Computer Vision}, pages 549--566. Springer, 2022.

\bibitem{yan2020learning}
Y.~Yan, J.~Qin, J.~Chen, L.~Liu, F.~Zhu, Y.~Tai, and L.~Shao.
\newblock Learning multi-granular hypergraphs for video-based person re-identification.
\newblock In {\em Proceedings of the IEEE/CVF conference on computer vision and pattern recognition}, pages 2899--2908, 2020.

\bibitem{yin2019towards}
B.~Yin, L.~Tran, H.~Li, X.~Shen, and X.~Liu.
\newblock Towards interpretable face recognition.
\newblock In {\em Proceedings of the IEEE/CVF International Conference on Computer Vision}, pages 9348--9357, 2019.

\bibitem{zhang2020multi}
Z.~Zhang, C.~Lan, W.~Zeng, and Z.~Chen.
\newblock Multi-granularity reference-aided attentive feature aggregation for video-based person re-identification.
\newblock In {\em Proceedings of the IEEE/CVF conference on computer vision and pattern recognition}, pages 10407--10416, 2020.

\bibitem{zheng2015scalable}
L.~Zheng, L.~Shen, L.~Tian, S.~Wang, J.~Wang, and Q.~Tian.
\newblock Scalable person re-identification: A benchmark.
\newblock In {\em Proceedings of the IEEE international conference on computer vision}, pages 1116--1124, 2015.

\bibitem{zhu2024sharc}
H.~Zhu, W.~Zheng, Z.~Zheng, and R.~Nevatia.
\newblock Sharc: Shape and appearance recognition for person identification in-the-wild.
\newblock In {\em Proceedings of the IEEE/CVF Winter Conference on Applications of Computer Vision}, pages 6290--6300, 2024.

\bibitem{zhu2022pass}
K.~Zhu, H.~Guo, T.~Yan, Y.~Zhu, J.~Wang, and M.~Tang.
\newblock Pass:part-aware self-supervised pre-training for person re-identification.
\newblock In {\em European conference on computer vision}, pages 198--214. Springer, 2022.

\end{thebibliography}
}

\end{document}